# Intelligent Materials Modelling: Large Language Models Versus Partial Least Squares Regression for Predicting Polysulfone Membrane Mechanical Performance

Dingding Cao[a], Mieow Kee Chan[b,c]*, Wan Sieng Yeo[d], Said Bey[e], Alberto Figoli[f]

[a]School of Big Data, Baoshan University, Baoshan, 678000, Yunnan Province, China.
[b]Centre for Water Research, Faculty of Engineering, Built Environment and Information Technology, SEGi University. Jalan Teknologi, Kota Damansara, 47810 Petaling Jaya, Selangor Darul Ehsan, Malaysia.
[c]School of Energy and Chemical Engineering, Xiamen University Malaysia, Jalan Sunsuria, Bandar Sunsuria, 43900 Sepang, Selangor, Malaysia.
[d]Oil and Gas Engineering Department, Faculty of Engineering, Universiti Malaysia Sabah, Jalan UMS, 88400 Kota Kinabalu, Sabah, Malaysia.
[e]Laboratory of Membrane Processes and Separation and Recovery Techniques, Faculty of Technology, University of Bejaia, Algeria.
[f]Institute on Membrane Technology (ITM-CNR), Via P.Bucci 17/C,87030 Rende (CS)- Italy.

**Abstract**

Predicting the mechanical properties of polysulfone (PSF) membranes from structural descriptors remains challenging due to extreme data scarcity typical of experimental studies. To investigate this issue, this study benchmarked knowledge-driven inference using four large language models (LLMs) (DeepSeek-V3, DeepSeek-R1, ChatGPT-4o, and GPT-5) against partial least squares (PLS) regression for predicting Young's modulus (E), tensile strength (TS), and elongation at break (EL) based on pore diameter (PD), contact angle (CA), thickness (T), and porosity (P) measurements. These knowledge-driven approaches demonstrated property-specific advantages over the chemometric baseline. For EL, LLMs achieved statistically significant improvements, with DeepSeek-R1 and GPT-5 delivering 40.5% and 40.3% of Root Mean Square Error reductions, respectively, reducing mean absolute errors from 11.63±5.34% to 5.18±0.17%. Run-to-run variability was markedly compressed for LLMs (≤3%) compared to PLS (up to 47%). E and TS predictions showed statistical parity between approaches ($q\geq0.05$), indicating sufficient performance of linear methods for properties with strong structure-property correlations. Error topology analysis revealed systematic regression-to-the-mean behavior dominated by data-regime effects rather than model-family limitations. These findings establish that LLMs excel for non-linear, constraint-sensitive properties under bootstrap instability, while PLS remains competitive for linear relationships requiring interpretable latent-variable decompositions. The demonstrated complementarity suggests hybrid architectures leveraging LLM-encoded knowledge within interpretable frameworks may optimise small-data materials discovery.

# 1. Introduction

Polysulfone (PSF) membranes are widely used for wastewater treatment, hemodialysis, and gas separation owing to their excellent thermal stability, chemical resistance, and tunable microstructure [1]. By manipulating the membrane formulation, such as incorporating additives and controlling the phase inversion process, the microstructure and surface properties of the membranes can be tailored to meet diverse application requirements [2]. Mechanical properties of a membrane, such as the Young's modulus (E), tensile strength (TS), and elongation at break (EL), are critical for a reliable and long term application. Studies have shown that microstructural parameters, such as pore size and porosity, as well as the composition of PSF membranes, have significant impacts on their macroscopic mechanical properties [3,4]. For instance, the presence of a small amount of additive, such as cellulose nanofibers, in the PSF membranes enhanced the yield stress from 3.81 MPa to 5.03 MPa, while the E was increased from 175.05 MPa to 234.5 MPa. Nevertheless, at a higher cellulose nanofiber loading of 0.5 wt.%, both the E and yield stress slightly decreased, due to nanofiber agglomeration, which affects the structural integrity. It is notable that the porosity of the PSF membrane also decreased with increasing cellulose nanofiber content, from 84% to 64% at the optimal loading of 0.3 wt.% [5].

A similar observation was reported when blending pentaerythritol-cored hyperbranched polyester additives (HBPEs-PER) into a PSF matrix. The presence of HBPEs-PER with a molecular weight of 2,160 g/mol in the PSF membrane promoted chain entanglement between PSF and hydroxyicosapentaenoic acids molecules, thereby increasing the breaking strength from 4.9 MPa to 6.6 MPa [6]. This indicates the complex interactive relationship between the porosity, polymer chain orientation, and additive compatibility in affecting the mechanical properties of a membrane. The relationships are highly nonlinear and interdependent [7], thus a simple linear correlation could not fully describe this complex structure–property interaction. Hence, there is a need to develop an advanced predictive model to tailor and optimise the PSF membrane formulation for practical application.

The major obstacle to developing accurate structure–property predictive models is the limited number of experimental data [8]. This is due to the time-consuming membrane fabrication process, labour-intensive characterisation tests, and the use of costly reagents to quantify the membrane separation performance. Consequently, only limited data were reported in the membrane experimental research papers, and typical dataset sizes were less than a dozen samples [9]. According to a recent review, small data is the main challenge faced by materials scientists in the effort to accelerate the new and sustainable materials discovery, and handling complex and high-dimensional data. In material science research, the fine-grained datasets were obtained from experiments conducted by experienced researchers, and thus the data were usually limited in quantity but high in quality [10,11]. These small datasets were produced from the well-controlled laboratory conditions using the well-established and globally-recognised scientific methods. Hence, small datasets tend to exhibit lower noise and fewer confounding factors, thereby enabling a clearer elucidation of causal relationships between material characteristics and properties [10]. Nevertheless, small datasets can lead to imbalanced training sets and significantly increase the probability of overfitting or underfitting [12]. The challenge is intensified when the data points are extremely sparse, where the training samples may not be available at the critical regions. This results in a highly uncertain and unreliable model prediction [13]. Therefore, in the context of materials science research with limited data, there is an urgent need to develop modelling strategies that can effectively extract knowledge from sparse observations and datasets, while preserving the physical meaning and generalizability, thereby maximising the use of the information contained in small sample sets.

Partial least squares (PLS) regression is a classical chemometric method that has been widely used for material property prediction because of its ability to cope with multicollinearity and small sample sizes [14]. PLS works by extracting latent components that represent joint features of the predictors and responses, projecting high-dimensional descriptors onto a low-dimensional space to maximize covariance with the target variables. This allows the model to achieve a robust linear regression even when the number of descriptors exceeds the number of samples [15]. PLS is widely adopted in polymer membrane research for analysing small datasets, due to its simple model form and straightforward interpretability. However, its inherent linear mapping restricts its predictive performance in modelling the strong nonlinear structure–property relationships. Additionally, PLS is highly sensitive to individual outliers, particularly when the sample size is extremely small. Thus, the selection of model parameters, such as the number of latent variables, becomes unstable and adversely affects the prediction reliability [16]. Nevertheless, in the absence of better alternatives, PLS remains a useful baseline. A PLS baseline model for predicting membrane mechanical properties was established in this study and subsequently used as the benchmark for comparative evaluations of newly proposed methods.

In recent years, the breakthroughs in artificial intelligence have led to the emergence of large language models (LLMs) as a novel approach for small-data modelling [17]. Transformer-based LLMs such as GPT-4 and GPT-5 are pre-trained on massive text corpora and thus exhibit powerful natural language understanding and reasoning capabilities. Thus, LLMs can accomplish complex tasks without the need for explicit programming, including writing code and answering domain-specific questions [18]. In material science research, LLMs can manage and link complex information while effectively handling ambiguity, particularly when the data are sparse and inconsistent. Their accuracy is highly reliable when they are embedded with external simulation and (or) human-in-the-loop validation. In the recent publication by Bran *et al.* [19], LLM was deployed for a chromophore case study. It involved automated literature mining, data extraction, experimental design assistance, and preliminary data analysis [20]. The selected molecule was synthesised and experimentally validated by the researchers. This confirmed the reliability of LLM-assisted decision-making. More importantly, the study demonstrated that LLMs can be fine-tuned to a specific domain, which is a significant step-change in materials discovery [21]. By learning from large volumes of scientific literature, LLMs serve as implicit knowledge bases, integrating cross-domain background information into small-data tasks [18].

In an ultrafiltration membrane study conducted by Gao *et al.* [22], tree-based machine learning models, including XGBoost and CatBoost, were trained on 320 literature data points to correlate key fabrication parameters, for instance, additive loading, polymer concentration, pore-former molecular weight, and pore-former content, together with operating conditions to estimate the membrane separation performance and antifouling properties. The optimized models achieved a Coefficient of Determination ($R^2$) of 0.83 with a Root Mean Square Error (RMSE) of 68.24 L/($m^2$.h.bar) for water permeability, and a $R^2$ of 0.84 with a RMSE of 6.60 % for bovine serum albumin removal. However, these machine learning models typically rely on hand-crafted numerical features and moderately sized datasets, and their performance degrades when the sample size is extremely limited. In contrast, LLMs are well trained on large corpora of scientific knowledge that can be transferred as prior information to fill gaps when they are applied to address problems which involved sparse and (or) limited new data [17, 21]. Furthermore, the attention-based reasoning mechanism of LLMs offers a degree of interpretability in the model's decision-making process. This is an additional benefit for scientific research. By harnessing the knowledge-driven inference capabilities of LLMs, particularly in the context of the small-data challenge in materials science research, LLMs improve the prediction accuracy while maintaining physical interpretability. In a polymer property prediction study, Gupta et al. [23] fine-tuned LLaMA-3-8B and GPT-3.5 on 11,740 polymer entries, which were represented as SMILES strings, to predict key polymer properties such as glass transition temperature ($T_g$), melting temperature ($T_m$), and decomposition temperature ($T_d$). It was carried out by using an instruction-style natural-language prompt, such as "If the SMILES of a polymer is <SMILES>, what is its $T_g$?" The result showed that the best single-task LLaMA-3 model achieved low RMSE values on the held-out test set. The RMSE values were 39.48 K for $T_g$, 58.23 K for $T_m$, and 77.11 K for $T_d$. This indicates a strong predictive performance of the LLaMA-3 model. This demonstrates that LLMs can learn meaningful structure–property mappings directly from text-encoded molecular representations, bypassing the labour-intensive feature engineering required by conventional machine learning pipelines.

In this study, LLMs, including the DeepSeek series and OpenAI's GPT series, are introduced for small-sample prediction of the mechanical properties of PSF membranes. The predictive performance of the selected LLMs was compared to the traditional linear PLS approach. Through carefully designed prompts, each sample's structural parameters were embedded in concise, scientific natural-language descriptions and served as the inputs to multiple LLMs for prediction. Under the same leakage-free leave-one-out cross-validation protocol used for PLS, the predictive performance of both PLS and LLMs approaches was compared under tightly controlled and fair experimental conditions. This study explores the potential of LLMs as a knowledge-driven tool to estimate membrane mechanical strength using limited physical properties data. The comparison of LLMs with PLS reveals the ability of models to capture the nonlinear structure-performance relationships under small-data conditions. The findings provide insights for membranologists to design preliminary membrane formulation and complement the conventional chemometric modelling.

## 2. Materials and Methods

### 2.1 Dataset Collection and Description

Ten PSF membranes labelled as S1–S10 in **Table 1**, were prepared from 20 wt% PSF, 15 wt% polyvinylpyrrolidone in dimethylacetamide under varied coaxial coagulant and its velocity [24]. The data were carefully selected by experienced membranologists, after considering the effect of the coagulant on membrane physical properties, including pore diameter (PD), contact angle (CA), thickness (T), and porosity (P), on the mechanical strength. Notably, the experimental data were measured according to the well-recognised laboratory methods.

The PD of the selected ten PSF membranes ranged from 0.298 to 0.842 μm, with CA from 66.4° to 94.6°, indicating a good coverage of both hydrophilic and hydrophobic behaviour of PSF membrane, as shown in **Table 1**. The membrane T fall within 0.136 to 0.273 mm, with P ranging from 69.09 to 79.16%. These four parameters were labelled as structural descriptors. The corresponding mechanical properties, such as E, TS and EL, are within 90.18–235.65 $N/mm^2$, 3.97 to 9.65 $N/mm^2$, and 42.07 to 69.04%, respectively.

**Table 1**. Polysulfone membrane dataset characteristics [24].

| Sample | PD (μm) | CA (°) | T (mm) | P (%) | E ($N/mm^2$) | TS ($N/mm^2$) | EL (%) |
|---|---|---|---|---|---|---|---|
| S1 | 0.522 | 94.6 | 0.273 | 77.67 | 117.17 | 4.82 | 42.07 |
| S2 | 0.364 | 93.3 | 0.202 | 79.16 | 90.18 | 3.97 | 46.61 |
| S3 | 0.569 | 78.9 | 0.175 | 78.87 | 152.50 | 6.56 | 49.13 |
| S4 | 0.451 | 66.4 | 0.136 | 73.60 | 231.78 | 9.61 | 61.30 |
| S5 | 0.408 | 66.6 | 0.154 | 73.49 | 182.59 | 7.43 | 60.69 |
| S6 | 0.336 | 79.5 | 0.177 | 69.09 | 235.65 | 9.65 | 64.46 |
| S7 | 0.403 | 81.3 | 0.240 | 71.18 | 176.51 | 7.25 | 69.04 |
| S8 | 0.319 | 84.5 | 0.180 | 78.32 | 126.42 | 5.22 | 42.91 |
| S9 | 0.842 | 90.0 | 0.224 | 78.86 | 107.67 | 4.53 | 51.25 |
| S10 | 0.298 | 82.9 | 0.188 | 74.02 | 168.57 | 7.13 | 67.15 |

## 2.2 Modelling Framework

**Figure 1** illustrates the design of the modelling experiment. The evaluation framework adopted a leave-one-out cross-validation (LOOCV; n = 10) scheme with two parallel branches executed within each fold. A data-driven branch applying bootstrap-perturbed PLS with inner LOOCV for latent-variable selection, and a knowledge-driven branch supplying the nine training samples as structured natural-language prompts to four LLMs (DeepSeek-V3, DeepSeek-R1, ChatGPT-4o, and GPT-5 [25-28]) under closed-book conditions. The entire pipeline is repeated five times with independent seeds, yielding fifty out-of-fold predictions per method that are subsequently evaluated via RMSE, Mean Absolute Error (MAE), $R^2$, and paired statistical tests.

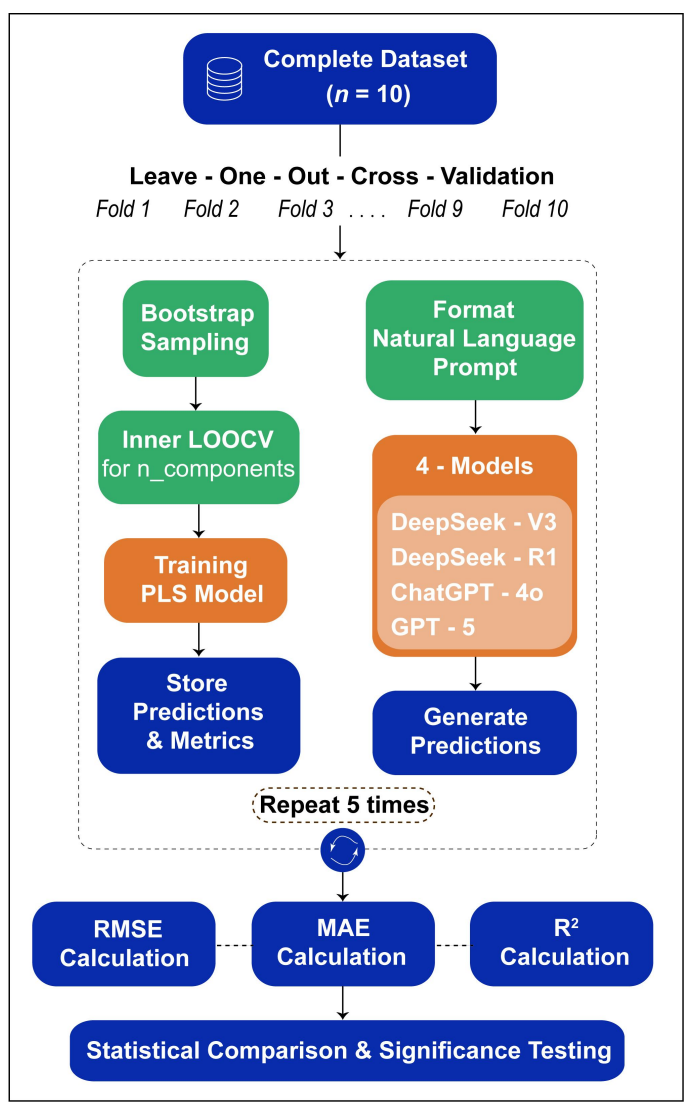

**Figure 1**. Cross-validated evaluation pipeline comparing the data-driven PLS baseline with knowledge-driven LLMs under LOOCV (n=10) with five repeated runs per method.

### 2.2.1 Data-Driven Baseline using Standardised PLS and Inner-LOOCV

A chemometric baseline was established using single-response PLS to map the four structural descriptors to the mechanical properties [29]. Model assessment was conducted using an outer LOOCV (n = 10) scheme. In each outer fold, all experimental data pre-processing steps and model fitting were confined to the training subset to prevent train–test leakage. Then, the held-out sample was transformed with the training statistics and predicted exactly once. To quantify run-level variability under extreme data scarcity, the training subset in each outer fold was perturbed by bootstrap resampling (with replacement, fixed size) before model selection and refitting. The complete fold-wise workflow was repeated a minimum of five times with independent seeds.

Predictors, $\widehat{X}$ as shown in **Eq. (1)** were standardised within each outer fold by z-scoring the training subset and applying the same transformation to the held-out sample, while targets, y, is unscaled and remained in physical units to preserve interpretability [29].

$$\widehat{X} = \frac{X - \mu_X}{\sigma_X} \qquad (1)$$

Here, $\mu_X$ and $\sigma_X$ denote the mean and standard deviation (SD) of each descriptor computed on the (bootstrapped) training data within the outer fold, respectively, PLS was fitted separately for each target. With $\widetilde{X} \in R^{n \times p}$ and $y \in R^n$, latent scores $t_a = \widetilde{X} w_a$ were extracted to maximise covariance with $y$, where $w_a$ are the x-weights, and regression was performed in the latent space. Predictions in the original descriptor space are shown in **Eq. (2)** [29]:

$$\hat{y} = \widetilde{X} B + \varepsilon \qquad (2)$$

where $B$ is the regression matrix estimated on the training subset.

The number of latent variables $n_{components}$ was chosen inside each outer fold by inner LOOCV conducted on the bootstrapped training data. Candidate dimensions $k = 1, \ldots, \min(p, n_{tr}-1)$ were evaluated by minimizing inner-fold validation error [30]. The selected $k$ was then refit on the full bootstrapped training subset and used to predict the single held-out sample. This procedure yielded one out-of-fold prediction per sample per run (ten predictions per run, five runs per target), forming the leakage-free PLS baseline for all subsequent comparisons.

### 2.2.2 Knowledge-Driven LLMs and Prompting

Knowledge-driven inference was performed by using four LLMs, namely DeepSeek-V3, DeepSeek-R1, ChatGPT-4o, and GPT-5. These models were selected to span complementary architectures and reasoning regimes, as shown in **Table 2**. All evaluations were conducted in a closed-book, text-only setting under a single standardised prompt; external tools, retrieval, and code execution were disabled. This heterogeneity in model design was intended to maximise complementary strengths while enabling a controlled comparison under identical inference constraints.

Prompting was done by following a compact, iterative strategy guided by three principles: sufficiency of scientific context, strict output determinism, and isolation of supervision. An initial parameter-list template (comma-separated descriptors) produced unstable numeric formatting and occasional off-schema strings. It was therefore replaced by a context-enriched template in which each descriptor was embedded in a concise scientific sentence, situating numbers within domain semantics while preserving precision. A format-standardised template was then finalised, comprising: (i) brief task framing and constraints, (ii) a fold-specific reference table containing the nine non-held-out specimens with descriptors and targets, and (iii) explicit output requirements. Responses were constrained to a single CSV with schema model_name, run, sample, property, units, predicted, exactly three rows per target, verbatim unit strings, and two-decimal rounding (Any non-numeric prose was disallowed to ensure deterministic parsing).

Within each outer LOOCV fold, only the nine reference rows were visible as supervised context, whereas the held-out specimen was presented solely by its four descriptors. Sessions were isolated to prevent conversational carryover and cross-fold leakage. To quantify repetition-level variability under fixed prompts, five independent runs per fold were executed for every model without altering prompt content or decoding settings. The resulting CSVs were ingested directly for metric computation in **Section 2.5**.

Formally, for model *m* and run *r* in fold *i*, predictions were obtained by **Eq. (3)** [31].

$$\hat{y}_{m,r}^{(i)} = f_m(R^{(i)}, x_{target}^{(i)}; \pi) \qquad (3)$$

where $R^{(i)}$ denotes the fold-specific reference table (descriptors and targets for the nine training specimens), $x_{target}^{(i)}$ is the descriptor vector of the held-out specimen, and $\pi$ is the locked prompt template.

**Table 2.** Characteristics of selected LLMs used for knowledge-driven prediction.

| Model (release) | Provider (release date) | Architecture/ type | Context window (in/out tokens) | Reasoning | Weights/ license |
|---|---|---|---|---|---|
| DeepSeek-V3-0324 | DeepSeek (2024-12) | Mixture-of-Experts (MoE); large-scale; multi-token prediction | up to 128k | General LLM | Open weights; MIT |
| DeepSeek-R1-0528 | DeepSeek (2025-05) | RL-trained reasoning LLM | up to 128k | Native reasoning | Open weights; MIT |
| ChatGPT-4o (2024-05) | OpenAI (2024-05) | End-to-end multimodal LLM | 128k / ~16k | General LLM | Proprietary |
| GPT-5 (2025-08) | OpenAI (2025-08) | Flagship text-vision LLM; tool-use, and long-context | 400k / 128k | Adaptive reasoning | Proprietary |

### 2.2.3 Validation Protocol and Study Design

A rigorously controlled, leakage-free evaluation was implemented using an outer LOOCV (n=10) design. Let $D = \{(x_i, y_i)\}_{i=1}^{10}$ denotes the descriptor–target pairs, with $D = (E_i, TS_i, EL_i)$. For outer fold $i$, the training set is $D_{-i} = D \setminus \{(x_i, y_i)\}$ and the test item is $(x_i, y_i)$. All pre-processing, model selection, and fitting were executed strictly within $D_{-i}$. The held-out item was transformed by training statistics and predicted exactly once per run.

On the data-driven branch including PLS, bootstrap perturbations were applied to quantify run-level variability under extreme data scarcity [32]. For run $r \in \{1,...,5\}$ within fold $i$, a resample $B^{(i,r)}$ of size $|D_{-i}|$ was drawn with replacement from $D_{-i}$. Predictors in $B^{(i,r)}$ were standardised (z-scored), and the same transformation was applied to $x_i$. The number of latent variables, $k^{*(i,r)}$ was selected by inner LOOCV on $B^{(i,r)}$ as shown in **Eq. (4)** [30]:

$$k^{*(i,r)} = \arg\min_{k \in \{1,\ldots \min\{p, |D_{-i}|-1\}\}} RMSE_{inner}(k) \qquad (4)$$

After which, a PLS model with $k^{*(i,r)}$ components were refit on $B^{(i,r)}$ and used to generate out-of-fold predictions $\hat{y}_{PLS}^{(i,r)}$ for the single hold-out. This yielded ten out-of-fold predictions per run and five runs per target, forming the PLS baseline for subsequent comparisons.

On the knowledge-driven branch (LLMs), inference was conducted with DeepSeek-V3, DeepSeek-R1, ChatGPT-4o, and GPT-5 under a closed-book, text-only regime using a locked, fold-specific prompt template (Section 2.2.2). Within fold $i$, only the nine training rows (descriptors and targets) were revealed as supervised context, while the held-out specimen was presented solely by its four descriptors. Sessions were isolated to prevent conversational carryover across folds. For each model $m$ and run $r \in \{1,...,5\}$, predictions were obtained as **Eq. (5)** [31]:

$$\hat{y}_m^{(i,r)} = f_m(R^{(i)}, x_i, \pi) \qquad (5)$$

where $R^i$ is the fold-specific reference table, and $\pi$ is the fixed prompt. Decoding parameters and output schema were held constant across repeats; external tools, retrieval, and code execution were disabled. All exact prompts for 10 folds * 5 runs per model and the corresponding raw CSV outputs are provided in the Supplementary Materials.

For each method and target, the out-of-fold residuals were assembled as **Eq. (6)** [33]:

$$e_{method,t}^{(i,r)} = \hat{y}_{method,t}^{(i,r)} - y_{i,t} \quad t \in \{E, TS, EL\} \qquad (6)$$

These residuals served as the sole inputs to the performance metrics and statistical tests described in **Section 2.5**. No model ever had access to its own held-out target during training, selection, or prompting, ensuring strict out-of-sample estimation.

**2.2.4 Metrics and Statistical Testing**

Performance was evaluated exclusively from out-of-fold residuals. For sample $i \in \{1,...,10\}$, run $r \in \{1,...,5\}$ method, $m \in M = \{$PLS, DeepSeek-V3, DeepSeek-R1, ChatGPT-4o, GPT-5$\}$, and target $t \in \{E, TS, EL\}$, the residual was defined as **Eq. (7)** [33]:

$$e_{m,t}^{(i,r)} = \hat{y}_{m,t}^{(i,r)} - y_{i,t} \qquad (7)$$

Within each run, three metrics were computed over the ten held-out items and subsequently summarised across the five repeats (reported as $mean \pm SD$). Let $\bar{y}_t = \frac{1}{10} \sum_{i=1}^{10} y_{i,t}$. The error metrics (all in physical units) used in this study were RMSE, MAE, and $R^2$, which are presented in **Eqs. (8)** to **(11)**, respectively [33].

$$RMSE_t^{(r)} = \sqrt{\frac{1}{10} \sum_{i=1}^{10} \left(e_{method,t}^{(i,r)}\right)^2} \qquad (8)$$

$$\text{MAE}_t^{(r)} = \frac{1}{10} \sum_{i=1}^{10} |e_{\text{method},t}^{(i,r)}| \qquad (9)$$

$$R_t^{2(r)} = 1 - \frac{\sum_{i=1}^{10} \left(e_{\text{method},t}^{(i,r)}\right)^2}{\sum_{i=1}^{10} (y_{i,t} - \bar{y}_t)^2} \qquad (10)$$

Between-method comparisons were conducted on paired absolute errors to respect fold–run coupling. For each target, two-sided Wilcoxon signed-rank tests were applied to **Eq. (11)** [34], yielding $n = 50$ paired observations per contrast.

$$d_{m\text{-PLS},t}^{(i,r)} = |e_{m,t}^{(i,r)}| - |e_{\text{PLS},t}^{(i,r)}|, \quad i=1\ldots 10,\ r=1\ldots 5 \qquad (11)$$

Multiplicity across LLMs was controlled within each target using the Benjamini–Hochberg (BH) procedure (false discovery rate $q = 0.05$). Significance was annotated as $q < 0.05$, $q < 0.01$, and n.s. $q \geq 0.05$. All tests were pre-specified and two-sided, with no normality assumptions beyond symmetry of paired differences under the null.

Effect sizes relative to the PLS baseline were summarised on RMSE. For method $m$ and target $t$, the percentage improvement was calculated using **Eq. (12)**,

$$\Delta \text{RMSE}_{m,t} = 100 \times \left(1 - \frac{\text{RMSE}_{m,t}}{\text{RMSE}_{\text{PLS},t}}\right) \qquad (12)$$

where $RMSE_{m,t}$ denotes the run-averaged RMSE over the ten held-out items. Nonparametric 95% confidence intervals for $\Delta \text{RMSE}_{m,t}$ were obtained via paired bootstrap resampling of fold–run indices $(i,r)$ (10,000 replicates), recomputing both numerator and denominator on each resample. Intervals excluding zero were interpreted as statistically significant improvements.

Within each target, four paired contrasts (each LLM vs PLS) were FDR-controlled using BH at q=0.05. p-values are two-sided. Pair-wise linear associations were examined using the Pearson coefficient as displayed in **Eq. (13)**.

$$r_{xy} = \frac{\sum_{i=1}^{n} (x_i - \bar{x})(y_i - \bar{y})}{\sqrt{\sum_{i=1}^{n} (x_i - \bar{x})^2} \sqrt{\sum_{i=1}^{n} (y_i - \bar{y})^2}} \qquad (13)$$

## 2.5 Implementation and Reproducibility

Analyses were executed on Ubuntu 22.04 with Python 3.11.8 (NumPy 1.26.4, SciPy 1.11.4, pandas 2.2.2, scikit-learn 1.4.2, matplotlib 3.8.4). Randomness was controlled by NumPy's (Generator (PCG64)) with seeds $s_r \in \{42,43,44,45,46\}$ for $r=1...5$. The same seed governed bootstrap resampling and inner-fold model selection to ensure run-level reproducibility [35,36].

For PLS (sklearn.cross_decomposition.PLSRegression), within each outer LOOCV fold, predictors were z-scored on the training subset, targets kept in physical units, model dimension chosen by inner LOOCV on the bootstrapped training data, then refit and used once to predict the hold-out. LLM inference (DeepSeek-V3-0324, DeepSeek-R1-0528, ChatGPT-4o-202405, and GPT-5-202508) was closed-book, text-only, with a locked prompt. The dataset CSV, exact PLS scripts, and finalised prompts (membranes labelled as S1–S10) are provided in Supplementary Materials. No external tools beyond **Sections 2.2** to **2.5** were used.

## 3. Results and Discussion

### 3.1 Pearson Correlation Matrix

The Pearson correlation matrix between the four structural descriptors, namely PD, CA, T, P and the respective mechanical properties, including E, TS, and EL; is illustrated in **Figure 2**. It is notable that the structural descriptors are negatively correlated to the mechanical properties, with P showing the strongest negative correlation across all three mechanical properties. The correlation with E is r = - 0.86, while TS is r = -0.84, and EL is r = -0.86. It is followed by CA, where r = -0.79, r = -0.79, and r = -0.57 for E, TS, and EL, respectively. This relationship is supported by the work done by Arat et al. [4]. When the P of the PSF/polyvinylpyrrolidone (PVP) based membrane was 20%, the TS was ~60MPa. At 33% P, a reduction in TS was observed with TS equal to ~45 MPa. This is due to the presence of increased pore interconnectivity and large void volume fractions in the porous membranes, which eventually reduced the membrane strength.

Similarly, the CA of a membrane is closely related to the membrane P and mechanical strength. Ratri et al. [37] produced cellulose acetate membrane using varied concentrations of acetone in the coagulation bath. When the acetone concentration increased from 25% to 75%, the membrane P increased from 65% to 70%. This is due to the faster demixing rate during the phase inversion process. As a result, a reduction in TS was observed, from 5.5 MPa to 2.6 MPa. A decrease in CA by 7° was also recorded. Dias et al. [38] further confirmed this finding. By adding solvent into the coagulation bath, the PSF/Polyethersulfone (PES) /PVP membrane exhibited higher P and lower CA compared to the membrane produced without solvent in the coagulation bath. This also reflects a lower mechanical strength in the earlier membrane.

T also has an impact on the mechanical strength, as described in the E and TS equations, where the cross-sectional areas of membranes are taken into calculation. Thus, moderate negative correlations between T with both E and TS were determined, with r = -0.57 and r = -0.59, respectively. Lastly, the PD contributed only weak negative trends with r=-0.35, -0.35, -0.31 for E, TS, and EL, respectively. In short, CA and P contributed significant impacts to the mechanical properties of membranes, while T and PD demonstrated weak but directionally consistent impacts.

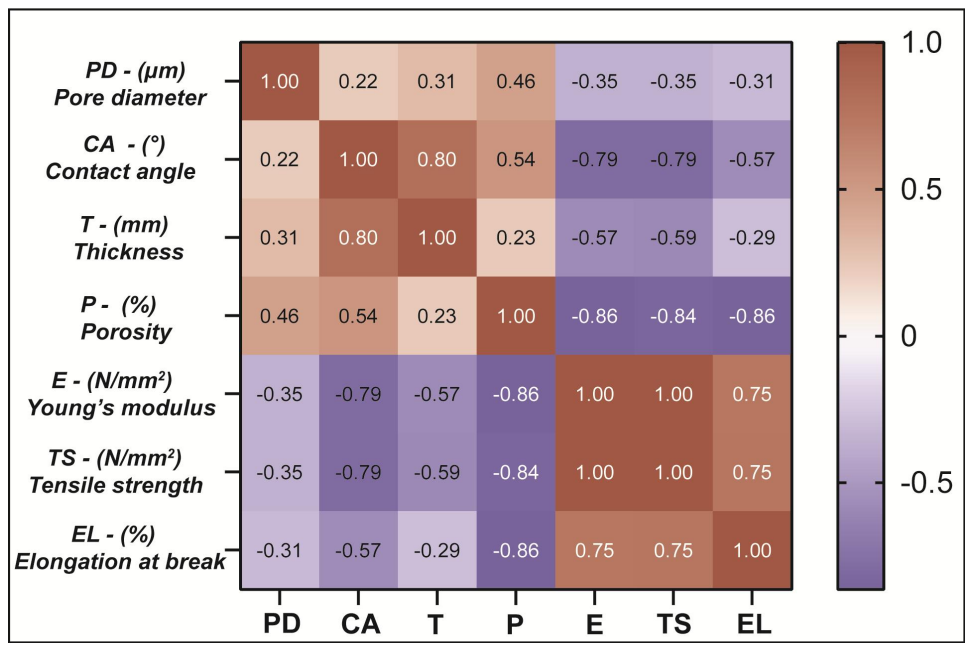

**Figure 2**. Pearson correlation matrix between structural descriptors (PD, CA, T, P) and mechanical properties (E, TS, EL) for the PSF membrane data set.

The mechanical properties data co-vary tightly, indicating a near one-dimensional response. E and TS are almost colinear, and both relate positively to EL with r = 0.748 and r = 0.750 [39,40]. This inter-target coherence implies that most variation in mechanical behaviour lies along a single principal axis, with limited orthogonal structure to discriminate among targets. The coherence of the dataset with the published data reflects the consistency of the observed membrane structural descriptors with the mechanical properties [24]. This supports the plausibility of the dataset as a reliable foundation for model development. The alignment between empirical measurement and theoretical expectations indicates that the data are not only statistically validated but also physically meaningful. Thus, the model developed from this reliable dataset provides a scientifically informed baseline for the comparative modelling that can be credibly interpreted.

### 3.2 Predictive Performance of the Models

**Figure 3** demonstrates the fold-wise distributions of absolute prediction errors for E, TS, and EL, respectively. The centre line in each boxplot represents the median value. The box represents the interquartile range (IQR), which spans from the 25th to the 75th percentile of the data. The whiskers represent data variability by extending to the most extreme values within 1.5 times the IQR, indicating the typical spread of the data outside the box. **Figures 3(a)** and **(b)** show that the medians and interquartile ranges of all four LLMs substantially overlap with the PLS baseline. Across both metrics, paired comparisons of each LLM against the PLS baseline yield two-sided Wilcoxon signed-rank tests with $q \geq 0.05$, indicating no statistically significant differences after BH FDR correction [41] . This indicates that the performances of all four LLMs are comparable to the PLS baseline in predicting the E and TS of the PSF membranes. No statistically significant difference between the PLS and any of the LLMs after BH FDR correction.

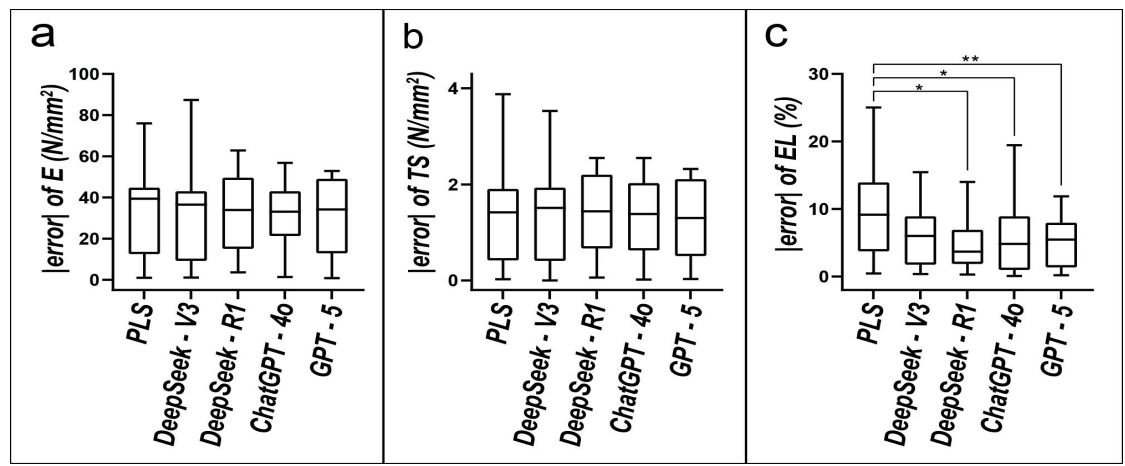

**Figure 3**. Distribution of absolute prediction errors for (a) E, (b) TS, and (c) EL, comparing PLS, DeepSeek-V3, DeepSeek-R1, ChatGPT-4o, and GPT-5. Centre line: median; box: IQR; whiskers: 1.5×IQR. Significance versus PLS (BH–FDR correction): *$q<0.05$, **$q<0.01$, n.s. $q\geq0.05$.

A distinct difference is observed in **Figure 3(c)**, where all four LLMs exhibit clearly smaller absolute errors compared to the PLS baseline in estimating EL. For the LLMs, the boxplots show absolute errors for EL below 10%, whereas the PLS model displays a wider spread, with the IQR range from 5% to 14 %. Pairwise comparisons versus PLS show *$q < 0.05$* (statistically significant) for ChatGPT-4o and DeepSeek-R1, and *$q < 0.01$* (highly statistically significant) for GPT-5, while DeepSeek-V3 does not differ significantly from PLS ($q \geq 0.05$). These results indicate that PLS and DeepSeek-V3 perform comparably for EL prediction, GPT-5 achieves the best performance among the LLMs, and DeepSeek-R1 and ChatGPT-4o are moderately better than PLS based on the observed error distributions. This finding is consistent with the data presented in **Table 3**. GPT-5 exhibited the smallest RMSE of $6.25 \pm 0.20$, the lowest MAE of $5.18 \pm 0.17$, and the highest $R^2$ of $0.58 \pm 0.03$ in predicting EL. Its performance is followed by ChatGPT-4o, DeepSeek-R1, DeepSeek-V3, and lastly PLS. This ranking is consistent with the substantially higher repeat-level variability observed for PLS, where the RMSE equals $16.00 \pm 10.03$. This confirmed the stability of LLMs in predicting the EL of PSF membranes.

Table 3. RMSE, MAE, and $R^2$ for predicting E, TS, and EL of PSF membranes using PLS and four LLMs..

| Models | E (N/mm²), [mean ± SD] | | | TS (N/mm²), [mean ± SD] | | | EL (%), [mean ± SD] | | |
|---|---|---|---|---|---|---|---|---|---|
| | RMSE | MAE | R² | RMSE | MAE | R² | RMSE | MAE | R² |
| **GPT-5** | 35.63 ± 0.53 | 30.84 ± 0.67 | 0.43 ± 0.02 | **1.50 ± 0.03** | 1.29 ± 0.04 | 0.38 ± 0.02 | **6.25 ± 0.20** | 5.18 ± 0.17 | 0.58 ± 0.03 |
| **ChatGPT-4o** | **34.58 ± 2.40** | 31.77 ± 2.47 | 0.46 ± 0.07 | 1.52 ± 0.11 | 1.32 ± 0.10 | 0.36 ± 0.09 | 7.73 ± 0.72 | 5.75 ± 0.47 | 0.36 ± 0.12 |
| **DeepSeek-R1** | 36.62 ± 0.57 | 32.35 ± 0.72 | 0.40 ± 0.02 | 1.59 ± 0.10 | 1.39 ± 0.14 | 0.30 ± 0.09 | 6.99 ± 2.74 | 5.51 ± 1.91 | 0.41 ± 0.38 |
| **DeepSeek-V3** | 40.81 ± 1.86 | 33.94 ± 1.86 | 0.25 ± 0.07 | 1.66 ± 0.07 | 1.39 ± 0.08 | 0.23 ± 0.06 | 8.77 ± 0.77 | 6.61 ± 0.43 | 0.17 ± 0.14 |
| **PLS** | 42.80 ± 5.34 | 35.51 ± 4.49 | 0.17 ± 0.21 | 1.80 ± 0.29 | 1.45 ± 0.26 | 0.08 ± 0.29 | 16.00 ± 10.03 | 11.63 ± 5.34 | -2.59 ± 4.73 |

Table 4. Relative RMSE improvement (%) of LLMs over the PLS baseline for E, TS, and EL, with corresponding CI.

| Model | E (%), [95% CI] | TS (%), [95% CI] | EL (%), [95% CI] |
|---|---|---|---|
| PLS | 0.0% [0.0, 0.0] | 0.0% [0.0, 0.0] | 0.0% [0.0, 0.0] |
| DeepSeek-V3 | -9.8% [-121.1, 40.9] | 0.2% [-101.0, 47.4] | 16.7% [-25.7, 52.8] |
| DeepSeek-R1 | 7.2% [-54.0, 42.2] | 10.2% [-43.5, 44.0] | **40.5% [11.4, 64.0]** |
| ChatGPT-4o | 7.1% [-72.2, 42.6] | 9.2% [-60.7, 48.7] | 26.5% [-7.4, 58.3] |
| GPT-5 | 3.5% [-69.7, 43.4] | 9.8% [-52.4, 48.6] | **40.3% [26.2, 49.8]** |

Table 5. Model rankings across E, TS, and EL, with overall average rank and per-run average rank (mean ± SD) based on RMSE-derived aggregation.

| Model | E Rank | TS Rank | EL Rank | Overall Avg. Rank | Avg. Rank per run (mean ± SD) |
|---|---|---|---|---|---|
| **GPT-5** | 2 | 1 | 1 | 1.33 | 1.47 ± 0.38 |
| **ChatGPT-4o** | 1 | 2 | 3 | 2.00 | 2.33 ± 0.94 |
| **DeepSeek-R1** | 3 | 3 | 2 | 2.67 | 2.80 ± 0.90 |
| **DeepSeek-V3** | 4 | 4 | 4 | 4.00 | 3.93 ± 0.55 |
| **PLS** | 5 | 5 | 5 | 5.00 | 4.47 ± 0.61 |

Notably, the MAE values of GPT-5 at 5.18, DeepSeek-R1 at 5.51, ChatGPT-4o at 5.75, and DeepSeek-V3 at 6.61 are substantially lower than PLS at 11.63 for EL, corresponding to ~55%, ~53%, ~51%, and ~43% MAE reductions, respectively. This indicates that GPT-5 and DeepSeek-R1 exhibited a better predictive performance compared to other LLMs and PLS. The level of statistical significance is further supported by the RMSE-based effect sizes and confidence intervals (CI) reported in **Table 4**, where PLS shows 0% improvement with a CI of [0.0, 0.0] for all properties, serving as the baseline. The improvements for EL are statistically significant for DeepSeek-R1 and GPT-5, as indicated by their positive effect sizes of 40.5% and 40.3%, respectively. Meanwhile, the corresponding 95% CIs are [11.4, 64.0] for DeepSeek-R1 and [26.2, 49.8] for GPT-5, where the ranges do not cross zero. This confirms the reliability of the observed performance improvements against sampling variability.

GPT-5 attains the smallest MAE (30.84) for E, followed by ChatGPT-4o (31.77), DeepSeek-R1 (32.35), DeepSeek-V3 (33.94), and PLS (35.51), as shown in **Table 3**. This indicates that LLMs perform slightly better compared to PLS, with MAE reductions of ~ 4% to 13% relative to PLS. For TS, MAEs are tightly clustered, with GPT-5 at 1.290, ChatGPT-4o at 1.320, DeepSeek-R1 at 1.386, DeepSeek-V3 at 1.387, and PLS at 1.450. Nevertheless, these modest improvements observed in LLMs compared to PLS may fall within the statistical uncertainty. This interpretation aligns with the overlapping box plots illustrated in **Figures 3(a)** and **3(b)**. In terms of RMSE-based effect sizes and CI as presented in **Table 4**, all LLMs show small and statistically uncertain changes for E and TS, as their 95% CIs include zero. This indicates that the predictive performance of LLMs is on par with the PLS approach for E and TS.

**Table 5** summarises the model rankings across EL, E, and TS, with overall average rank and per-run average rank according to the RMSE-derived aggregation. RMSE is selected as the reference metric because it is more sensitive to differences in model performance than MAE and $R^2$, as shown in Table 3. Variations across models are more distinct in RMSE, whereas MAE and $R^2$ exhibit smaller changes. The ranks for the mechanical properties, E, TS, and EL in **Table 5** are assigned based on the RMSE values in Table 3. The model with the smallest RMSE value is given a rank of 1, while the largest value is 5. The overall average rank is calculated by taking the mean of the three individual ranks for EL, E, and TS of a model. This shows an overall performance of a model in predicting EL, E, and TS. The result shows that GPT-5 achieves the best overall average rank at 1.33, followed by ChatGPT-4o at 2.00, DeepSeek-R1 at 2.67, DeepSeek-V3 at 4.00, and lastly, PLS at 5.00. A similar sequence is observed when the model performance is evaluated in terms of average rank per run. This was done by averaging the overall average rank across the five runs. This indicates that GPT-5 demonstrates the best predictive performance and also maintains consistent ranking stability across different runs, as shown by the smallest SD value.

## 3.3 Predictive Accuracy, Bias, and Uncertainty of LLMs and PLS Model

**Figure 4** presents the parity plots for E, where the predicted data by each model are compared with the corresponding experimental data. These scatter plots provide a clear visual assessment of model agreement, reflecting the degree to which the predictions align with the experimental data, as presented by the identity line in blue in **Figure 4**. In addition, system bias could be observed in the plots, to describe whether a model tends to consistently overestimate or underestimate E. In **Figures 4(a) – 4(e)**, the SD bars indicate run-to-run uncertainty, where a larger SD value represents greater model sensitivity to bootstrap perturbations in the training subset [32]. PLS model in **Figure 4(a)** shows large SD bars, particularly at S1, where the variability reaches ~ 44.86 N/mm², and at S10, where it is ~ 29.34 N/mm². Comparatively, the LLMs exhibit narrow SD bars, indicating a tighter repeatability across the five repeated runs. For instance, the variability falls within ~1–11 N/mm², as showed by DeepSeek-V3 at S1 with 2.68 N/mm² in Figure 4 (c) and GPT-5 at S1 with 3.17 N/mm² in **Figure 4(e)**. In **Figure 4(f)**, sample-wise patterns are consistent across all the models, with underestimation observed for S3, S4, and S6 and overestimation at S2, S5, and S8. Notably, a model-specific outlier appears at S9, where PLS overestimates the hold-out data by +84.91 N/mm², while all LLMs underestimate the data within the range of −24.17 to −7.85 N/mm². Comparatively, GPT-5 has the minimum run-to-run uncertainty as the SD bar is hardly observed in **Figure 4(e)**, indicating GPT as the best model in predicting E.

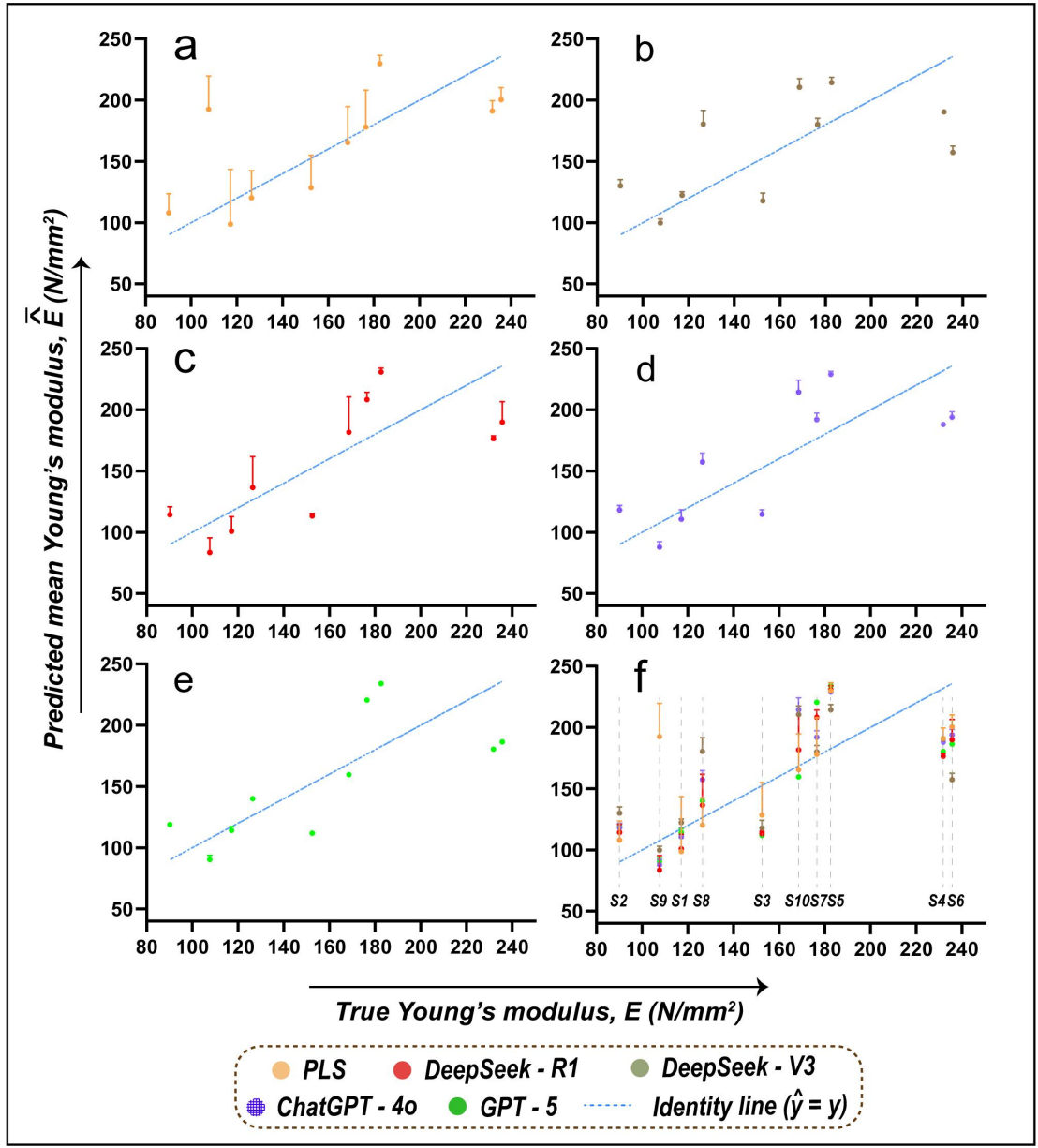

**Figure 4.** Parity plots for E using (a) PLS, (b) DeepSeek-R1, (c) DeepSeek-V3, (d) ChatGPT-4o, (e) GPT-5, and (f) all models combined.

**Figure 5** shows the parity plots of EL. Wide SD value per sample is observed in the PLS model, particularly for S3 and S6, as illustrated in **Figure 5(a)**. A large deviation from the identity line was also observed. This indicates high run-to-run uncertainty and weak model predictive stability. For instance, S6 shows the largest SD dispersion of 47.03%, while the SD is typically ≤ 3% for all LLMs. Additionally, mean signed biases remain small for all LLMs. DeepSeek-R1 and DeepSeek-V3 exhibit tighter clustering around the identity line, with smaller deviations and reduced uncertainty, as illustrated in **Figures 5(b)** and **5(c)**, respectively. ChatGPT-4o and GPT-5 display even closer alignment with the experimental EL data, with narrow SD bars and more consistent adherence to the expected trend. The combined plots presented by **Figure 5(f)**, low-EL sample, such as S1 and S8, are overestimated, and high-EL samples, such as S7 and S10, are underestimated. This well describes a clear regression-to-the-mean pattern, where the prediction tends to move toward the middle rather than capturing the full range of experimental data.

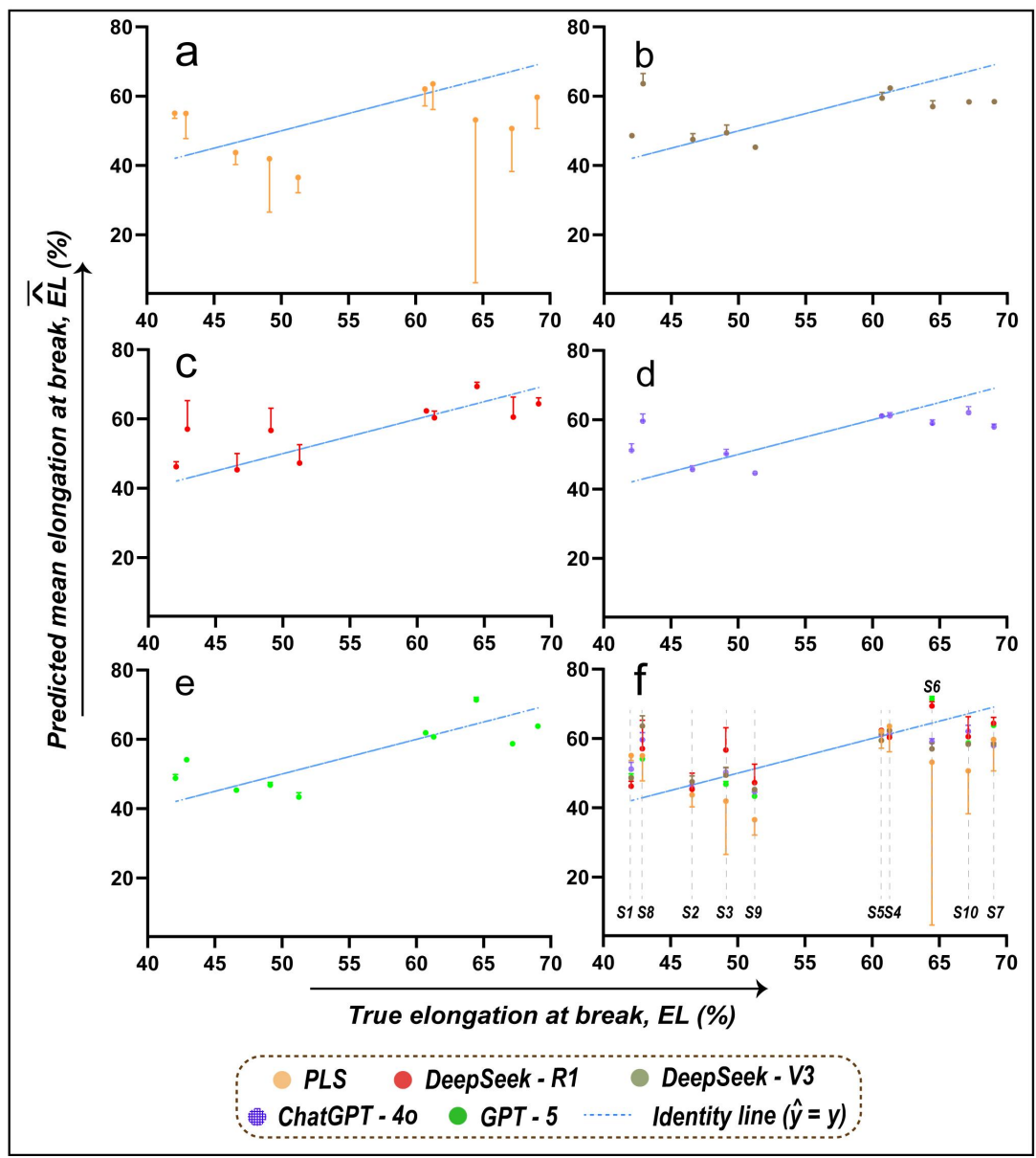

**Figure 5.** Parity plots for EL using (a) PLS, (b) DeepSeek-R1, (c) DeepSeek-V3, (d) ChatGPT-4o, (e) GPT-5, and (f) all models combined.

Parity plots for TS are presented in **Figure 6**. Similar to the plots in **Figures 4** and **5**, the SD bars of PLS model in **Figure 6(a)** are wider compared to the LLMs in **Figures 6(b)** to **6(e)**. This confirms the higher run-to-run uncertainty of the PLS model compared to all four LLMs. Notably, **Figure 6** shows a milder regression-to-the-mean behaviour in comparison to **Figures 4** and **5**, aligned well with the small MAE and RMSE data in **Table 3**. In the overall plots presented by **Figure 6(f)**, a systematic underestimation is observed for S3, S4, and S6, while S5 tends to be overestimated. A PLS specific deviation appears at S9, where PLS overestimates TS by +3.83 N/mm², in contrast to the LLMs, whose residuals remain close to zero, specifically within −1.30 to −0.36 N/mm². Similar to the analysis of **Figures 4** and **5**, GPT-5 exhibited very narrow SD bars across the 10 samples, indicating a stable model for TS prediction.

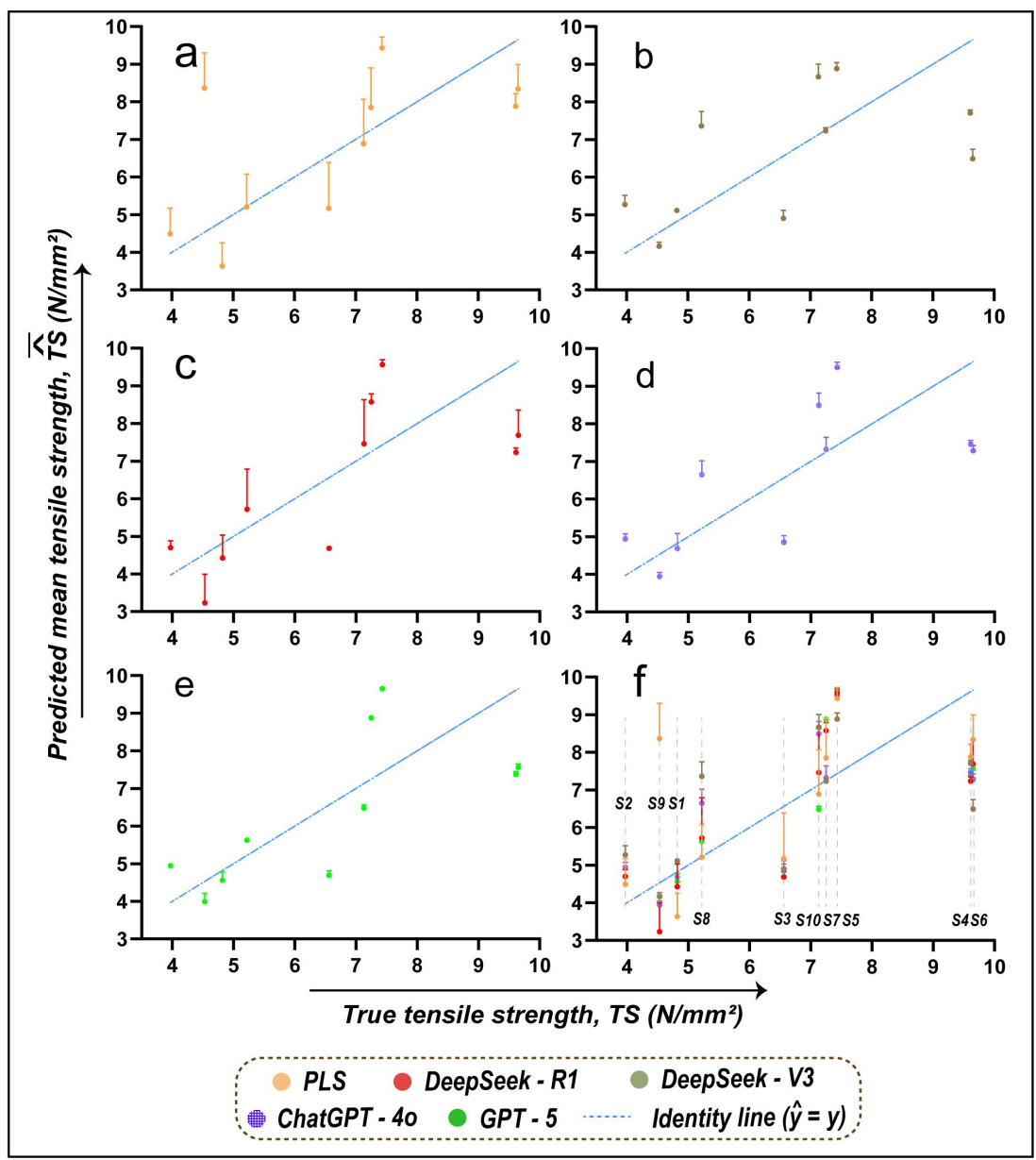

**Figure 6.** Parity plots for TS using (a) PLS, (b) DeepSeek-R1, (c) DeepSeek-V3, (d) ChatGPT-4o, (e) GPT-5, and (f) all models combined.

As a summary, the data distribution patterns observed in the parity plots showed in **Figures 4 - 6**, show that GPT-5 exhibited the tightest clustering around the identity line for E, TS, and EL, indicating a robust model in estimating the mechanical property of PSF membranes. It is followed by ChatGPT-4o, DeepSeek-R1,

DeepSeek-V3, and lastly PLS. This sequence aligns with the ranking results presented in **Table 5**, where models with lower RMSE and more stable performance demonstrate tighter clustering around the parity line and reduced bias. These findings also suggest that the predictive performances for E and TS are comparable between the PLS model and LLM, but the LLM is better in estimating EL. This is supported by the narrow SD bars for LLMs observed in **Figure 5(f)**, and low RMSE and MAE data of EL in **Table 3**, in relative to E and TS. Hence, it is reasonable to deduce that the LLMs, which is knowledge-encoded priors, enable more effective modelling of non-linear, constraint-aware deformation behaviour, even under conditions of extreme data scarcity, relative to PLS.

**3.4 Sample-Structured Residual Trends and Regression-to-the-Mean Behaviour**

Residual plots illustrated in **Figure 7** complement the parity plots in **Figures 4 to 6** by revealing sample-wise bias, magnitude-dependent error patterns, and model-specific error patterns that may not be visible in parity plots. For E, the residual plot in **Figure 7(a)** shows that the coherent, sample-level error patterns are observed across the models, where the same sample exhibits the same type of error. For instance, S3, S4, and S6 consistently underestimated, while S2, S5, and S8 are overestimated. This aligns with the observation in **Figure 4**. This indicates that most of the error is data-regime specific. A similar pattern is observed in TS, as shown in **Figure 7(b)**, where S3, S4, and S6 fall within the negative residual region, indicating underestimation; while S2, S5, and S8, fall within the positive residual region, indicating overestimation. Notably, for E, S9 is overestimated only by PLS, with a hold-out value of +84.9 N/mm², while all four LLMs underestimate S9 by ~ −24.17 to −7.85 N/mm². Similarly, the same pattern is observed in **Figure 7(b)** for TS, where PLS overestimated the hold-out by +3.84 N/mm² compared with the LLMs, while LLM residuals remained near zero, which is approximately -1.30 to - 0.36 N/mm². This indicates a model-specific error pattern, revealing the weakness of PLS in the model in comparison to LLMs.

Lastly, EL shows clear magnitude-dependent bias, as shown in **Figure 7(c)**. Low-EL samples such as S1 and S8 are consistently overestimated, whereas samples with high-EL values such as S7, S9, and S10 are underestimated across methods. For samples with mid-range EL values, ranging from ~47% - ~62%, the residuals cluster is near zero. This confirms that the magnitude depends on behaviour. In addition, the residual spread for the LLMs from the zero residual line, is visibly narrower than for PLS. This indicates smaller error amplitudes and more consistent repeat-level performance. The largest run-to-run dispersion is observed in the PLS baseline, particularly at S6. These patterns reflect the non-linear, constraint-sensitive nature of elongation and show that LLMs provide more stable predictions in a situation where the total number of experimental samples is extremely small.

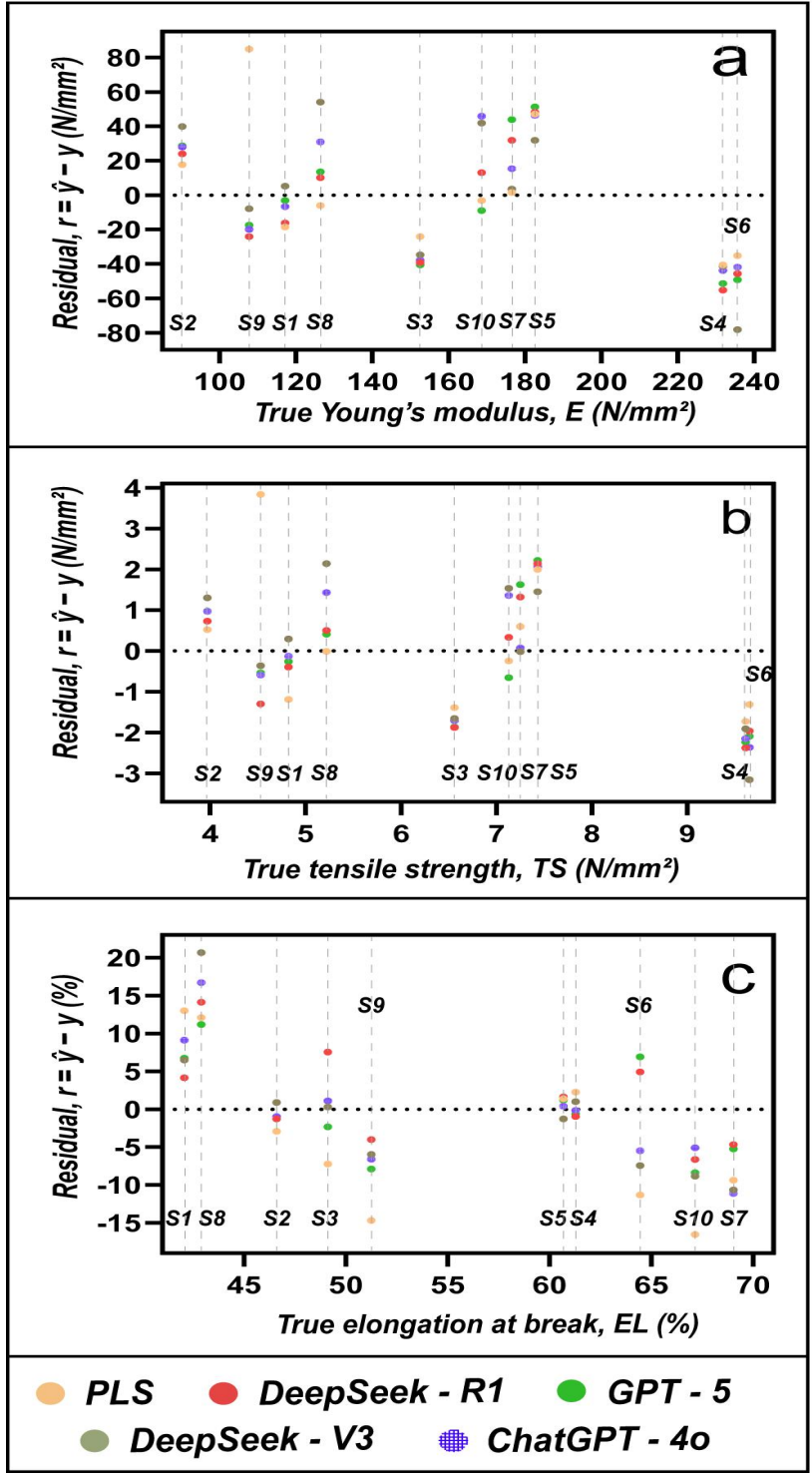

**Figure 7**. Residual plots for (a) E, (b) TS, and (c) EL across PLS and LLMs

## 3.5 Error Topology by Sample (Fold-wise Residual Structure)

Fold-wise residuals are compiled into sample–method heat maps in **Figure 8** to reveal the localized bias patterns under LOOCV. For E and TS, a stable topology emerges across all approaches where consistent underestimation is observed in S3, S4, and S6, while S2, S5, S7, S8, and S10 are overestimated. This can be clearly observed in **Figures 8(a)** and **8(b)**. This indicates a data-regime specific pattern, and this aligns with the pattern illustrated in the residual plot in **Figures 7(a)** and **7(b)**. In addition, an anomaly is identified at S9, where the PLS model exhibits a large positive residual, in contrast to uniformly negative residuals for the LLMs. This suggests a model specific error pattern, further confirming the analysis of the residual plot.

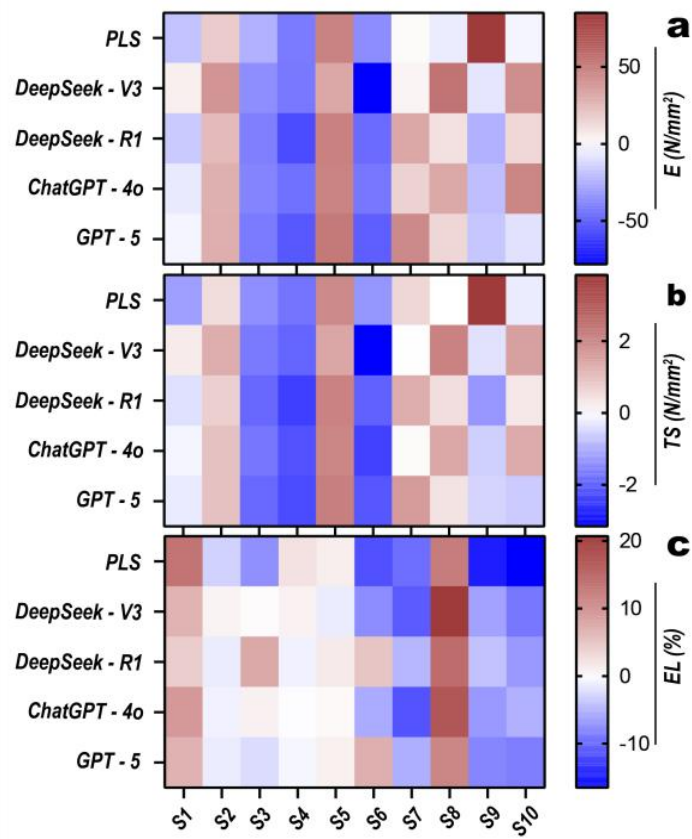

**Figure 8**. Fold-wise residual heat maps for (a) E (N/mm²), (b) TS (N/mm²), and (c) EL (%), with colour showing residual sign and magnitude, where blue denotes underestimation and red denotes overestimation.

Magnitude-dependent bias pattern is clearly seen in **Figure 8(c)**, where low-EL samples, including S1 and S8, are consistently overestimated while high-EL samples, namely S7, S9, and S10, are underestimated across all models. Despite the shared sign pattern, dispersion varies a lot between models, most notably at S6. LLMs stay close to the experimental data, but the PLS model sometimes gives very large errors. This indicates that knowledge-driven inference by LLMs reduces residual magnitude and stabilizes repeats precisely on the challenging folds that dominate EL error.

In short, the heat-map topology reveals data-regime–driven bias that is largely method-agnostic, punctuated by isolated, model-specific outliers, as shown by the PLS model at S9 for E and TS. The consistent sign pattern across models is in agreement with the regression-to-the-mean behaviour quantified in **Section 3.4**. The run-to-run dispersion for LLMs on EL is markedly smaller than for the PLS model, aligns with the patterns illustrated in the parity plots in **Figure 5**.

### 3.6 Limitations and Recommendations

The dataset is limited to the PSF membrane where the membrane P is manipulated by the composition of the coagulation bath and its temperature. Other membrane modification technique such as the addition of hydrophilic additives and surface copolymerisation, may affect the relationship between the studied parameters. The LLM prediction process was conducted under a fixed closed-book protocol. The sensitivity to prompt design, decoding parameters, or model versioning was assumed to remain consistent across all models. Future work should expand the dataset to determine the sample-size threshold at which data-driven methods regain superiority, incorporate additional physicochemical descriptors, and evaluate domain-informed prompt engineering strategies. This could encode materials physics constraints explicitly. Cross-system validation on membranes with different fabrication techniques, a wider range of mechanical properties, and varied application domains shall be conducted in the near future to confirm the model's applicability beyond PSF membranes and reduce risks of system-specific overfitting.

### 4. Conclusions

This study concluded that the LLMs complement the traditional chemometric approaches for predicting PSF membrane mechanical properties under extreme data scarcity. Using a leakage-free, closed-book protocol and rigorous leave-one-out validation, LLMs delivered a clear and statistically supported advantage for EL prediction. Both DeepSeek-R1 and GPT-5 demonstrated 40.5% and 40.3% RMSE reductions, respectively, and reduced mean absolute errors from 11.63±5.34% to 5.18±0.17%. Run-to-run variability showed by the parity plots, was markedly narrow for LLMs with SD≤3% compared to PLS, where the SD is up to 47%. Meanwhile, E and TS predictions remained statistically comparable across the models, where $q \geq 0.05$. This indicates that when structure–property relationships are essentially linear, which is consistent with first-order mechanical behaviour, linear covariance methods such as PLS perform adequately. LLMs should be prioritised for non-linear, constraint-sensitive properties prone to collinearity and instability under extreme data scarcity. These findings are in accordance with the residual plots and fold-wise residual heat map patterns in this study. Incorporating additional physicochemical descriptors, evaluating domain-informed prompt engineering strategies, and cross-system validations should be conducted in the future to confirm the model's applicability beyond PSF membranes and reduce risks of system-specific overfitting.

## 5. Data, Code, and Prompt Availability

All artifacts necessary for full reproducibility have been deposited in an openly accessible repository on GitHub at https://github.com/bsxyding/Intelligent-Materials-Modelling (commit a8d8aed).

## 6. Acknowledgments

Research Fundings supported by Universiti Malaysia Sabah (GKP 2501) and SEGi University (SEGiIRF/2025/FoEBEIT/4) are greatly acknowledged. Institutional support and access to instrumentation have been provided by Baoshan University.

## 7. Author Contributions

Contributions have been assigned according to the CRediT taxonomy. Conceptualization has been carried out by M. K. Chan and A. Figoli. Methodology has been developed by D. Cao, M. K. Chan, and W. S. Yeo. Software and data curation have been performed by D. Cao. Validation and formal analysis have been conducted by D. Cao and M. K. Chan. An investigation has been executed by D. Cao. Resources have been provided by M. K. Chan, S. Bey, and A. Figoli. Writing—original draft preparation has been undertaken by D. Cao, and M. K. Chan. Writing—review and editing have been completed by M. K. Chan, W. S. Yeo, S. Bey, and A. Figoli. Visualization has been prepared by D. Cao. Supervision and project administration have been provided by M. K. Chan and A. Figoli. The corresponding author is M. K. Chan. All authors have approved the final version of the manuscript.

## 8. Artificial Intelligence (AI) Transparency Statement

LLMs (DeepSeek-V3, DeepSeek-R1, ChatGPT-4o, and GPT-5) were employed as research objects for knowledge-driven prediction of membrane mechanical properties, as described in Sections 2.2.2 and 3. ChatGPT was additionally used to improve grammar, clarity, and readability of the manuscript. All AI-assisted text was reviewed, revised, and validated by the authors.

## 8. Competing Interests

No competing financial or non-financial interests have been declared.